\documentclass[10pt,twocolumn,letterpaper]{article}
\usepackage{iccv}

\usepackage[numbers, sort]{natbib}
\usepackage{times}
\usepackage{epsfig}
\usepackage{graphicx}
\usepackage{amsmath}
\usepackage{amssymb}
\usepackage{booktabs}
\usepackage{comment}
\usepackage{xcolor}
\usepackage{microtype}
\usepackage{enumitem}
\usepackage{multirow}
\usepackage{array}
\usepackage[symbol]{footmisc}

\usepackage[pagebackref=true,breaklinks=true,letterpaper=true,colorlinks,bookmarks=false]{hyperref}

\iccvfinalcopy 

\def\iccvPaperID{****} 

\ificcvfinal\fi

\begin{document}

\title{Unlimited Neighborhood Interaction for Heterogeneous Trajectory Prediction}
\author{Fang Zheng$^1$ ~Le Wang$^{2}$\footnote[1]{}  ~Sanping Zhou$^2$  ~Wei Tang$^3$ ~Zhenxing Niu$^4$  ~Nanning Zheng$^2$ ~Gang Hua$^5$ \\
$^{1}$School of Software Engineering, Xi'an Jiaotong University\\
$^{2}$Institute of Artificial Intelligence and Robotics, Xi'an Jiaotong University\\
$^{3}$University of Illinois at Chicago\\
$^{4}$School of Computer Science and Technology, Xidian University\\
$^{5}$Wormpex AI Research\\
}

\maketitle
\thispagestyle{empty} 
\footnotetext{$^*$Corresponding author.}
\begin{abstract}

    Understanding complex social interactions among agents is a key challenge for trajectory prediction.
	Most existing methods consider the interactions between pairwise traffic agents or in a local area, while the nature of interactions is unlimited, involving an uncertain number of agents and non-local areas simultaneously.
	Besides, they treat \emph{heterogeneous} traffic agents the same, namely those among agents of different categories, while neglecting people's diverse reaction patterns toward traffic agents in different categories. 
    To address these problems, we propose a simple yet effective Unlimited Neighborhood Interaction Network~(UNIN), which predicts trajectories of heterogeneous agents in multiple categories.
	Specifically, the proposed unlimited neighborhood interaction module generates the fused-features of all agents involved in an interaction simultaneously, which is adaptive to any number of agents and any range of interaction area.
	Meanwhile, a hierarchical graph attention module is proposed to obtain category-to-category interaction and agent-to-agent interaction. 
	Finally, parameters of a Gaussian Mixture Model are estimated for generating the future trajectories.
	Extensive experimental results on benchmark datasets demonstrate a significant performance improvement of our method over the state-of-the-art methods. 
\end{abstract}

\section{Introduction}

	\begin{figure}[t]
    \begin{center}
    \includegraphics[width=\linewidth]{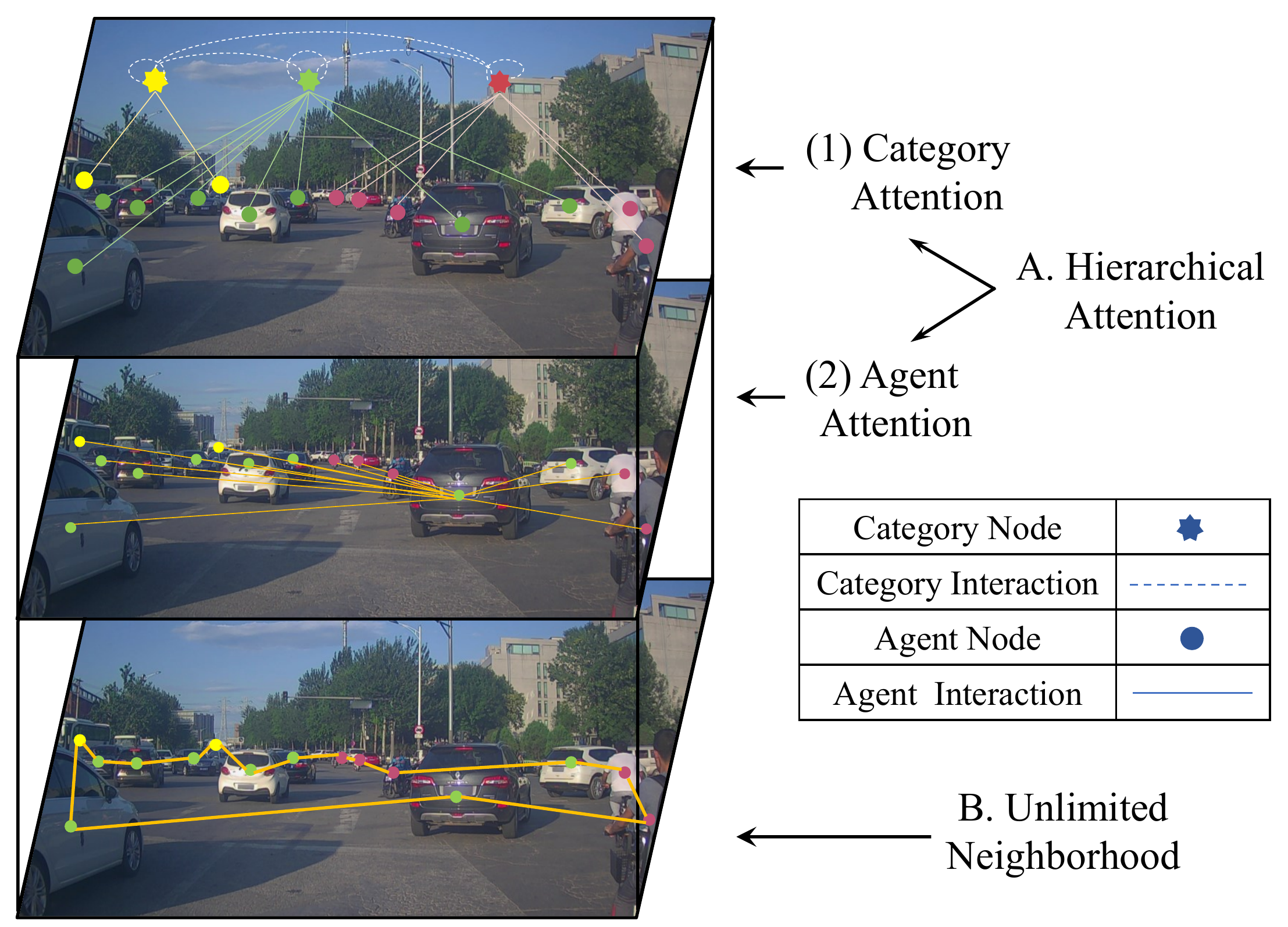}
    \end{center}
    \caption{Hierarchical Graph Attention \& Unlimited Neighborhood Interaction.
     Different marked shapes are used to distinguish the agent and category.
    \textbf{(A) Hierarchical Attention.} The agents marked by the same color belong to the same category. \textit{(1) Category Attention.} Every category interacts with each other and itself. Attention of one category to all categories (including the attention to itself) is transferred to all agents of the category. \textit{(2) Agent Attention.} We compute one's attention to the rest of the agents in the whole scenario, and the attention is directed. \textbf{(B) Unlimited Neighborhood.} We consider the interaction as among a 
   collective, rather than between two agents or in a small area. The behavior of any agent may influence a group of agents around the whole scenario.}
    \label{fig:start}
    \end{figure}
    
	The challenges hampering prediction accuracy largely 
	stem from the complex interactions among agents~\cite{alahi2016social,gupta2018social,zhang2019sr}. 
	Recent advances in this regard~\cite{alahi2016social,liang2020simaug,liang2020garden,bi2019joint,li2019interaction} mainly fall into two types: 
    Graph-based methods~\cite{mohamed2020social,9093456,yu2020spatio} build a spatial graph at each time step and aggregate the features from adjacent nodes;
    RNN-based methods~\cite{chandra2020forecasting,zhang2019sr,bisagno2018group} model each agent's trajectory with Recurrent Neural Networks~(RNNs) and pool hidden states within a surrounding area.
    
	However, these methods suffer from limitations. 
	%
	Graph-based methods~\cite{mohamed2020social} only exploit pairwise relation between the nodes,
	while other nodes are mixed and relayed. GCN with many layers suffers from over-smoothing problem~\cite{1NEURIPS2019_73bf6c41, 2018arXiv181005997K}.
	In contrast, the interaction in real-world traffic is much more complex than previously assumed, such as multilateral relations (relation among three or more agents).
	Namely, these methods are limited by inflexible numbers of interaction agents.
	
	Moreover, RNN-based methods~\cite{chandra2020forecasting,zhang2019sr,bisagno2018group} merely consider the local relations among an agent's manually defined surrounding area. 
	As a result, potential interaction participants outside of such ``surrounding area'' will be simply discarded.
	Namely, these methods are limited by such hand-crafted way for interaction agent selection.
	
\begin{figure}[t]
   \begin{center}
    \includegraphics[width=\linewidth]{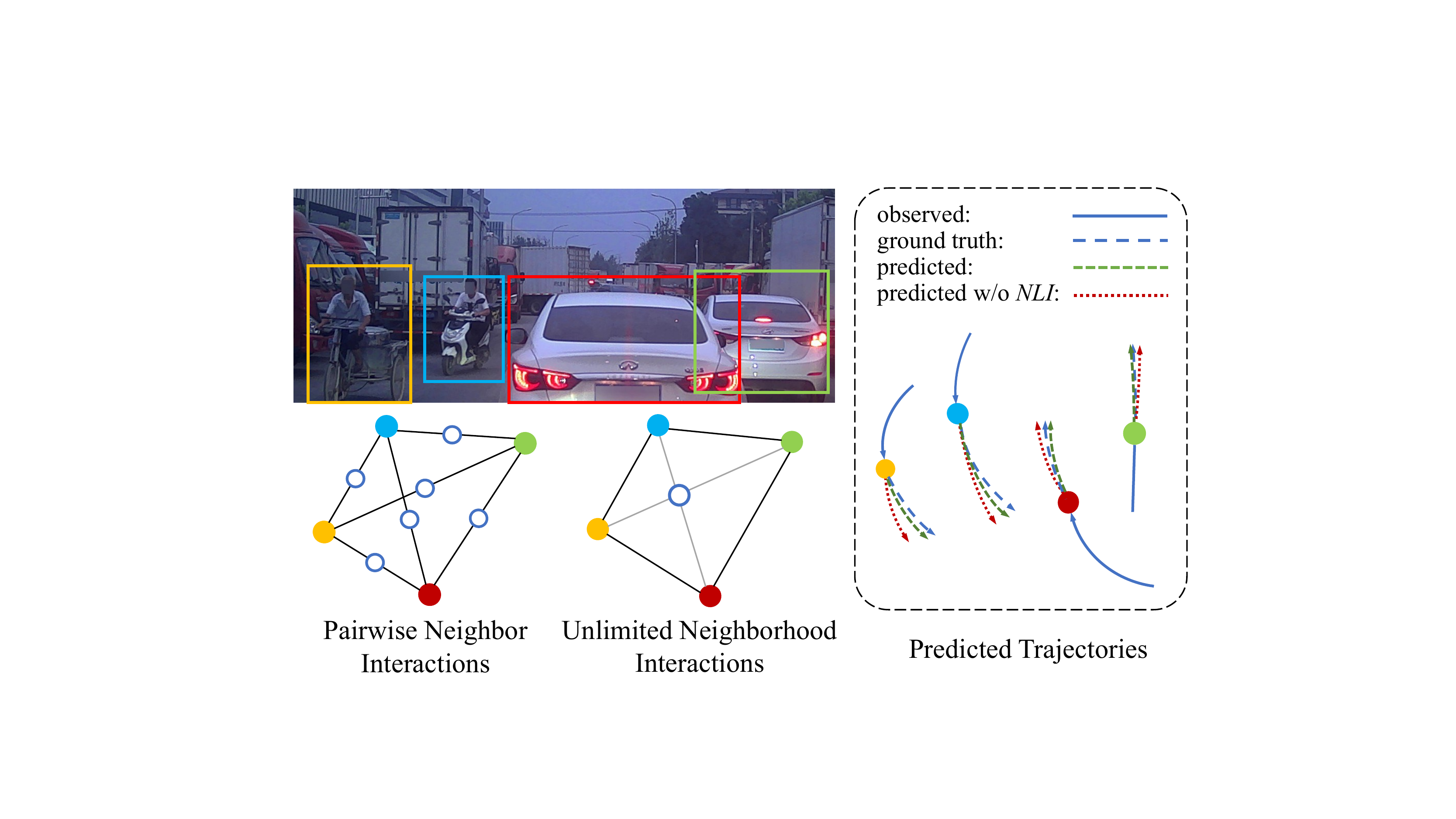}
     \end{center}
    \caption{Comparison between Unlimited Neighborhood Interaction and Pairwise Neighborhood Interaction (\eg, GCN). Different agents are enclosed in differently colored boxes, corresponding to solid circles in the same color. The hollow circle denotes an interaction. As can be seen, an interaction involves a group of agents in our method. On the right side, we show the predicted trajectories with or without our Unlimited Neighborhood.}
    \label{fig:nl}
\end{figure}

	To solve these problems, we propose the Unlimited Neighborhood on heterogeneous graph to predict the future trajectories of multi-categories~(\eg, pedestrians, bikes, cars, .etc), as shown in Figure~\ref{fig:start}.
	Unlimited Neighborhood means the interactions are not limited by the number of agents or the range of area. Namely, any agent in a scenario could be involved in an interaction, as illustrated in Figure~\ref{fig:nl}.
	In addition, many related works~\cite{gupta2018social,yu2020spatio} treat different agents as the homogeneous ones~(\ie, pedestrians), while a real traffic scenario usually involves heterogeneous agents~(\ie, agents in diverse categories).
	Due to the difference in movement patterns~(\eg, velocity, front and rear distance and the response to the interaction) for agents in different categories, trajectory prediction on heterogeneous agents is exactly more challenging compared to that on homogeneous ones.
	
	
    
	Specifically, we present a simple yet effective Unlimited Neighborhood Interaction Network for heterogeneous trajectory prediction, which models the hierarchical attention and fuses all agents involved in one interaction to predict the future trajectories for all agents with different categories simultaneously.
	Then, regarding the agents as nodes and the agents with the same category as a category node, we can construct a spatio-temporal-category graph combining spatial, temporal and category information together.
	The hierarchical graph attention module acquires the category-category attention and then the agent-agent attention on the constructed graph. 
	Note that the edges in the constructed graph are directed. Namely the edges are represented as a weighted asymmetric adjacency matrix to measure the interactions. 
	
	Once obtained the hierarchical interactions, an unlimited neighborhood interaction module is employed to capture the global information of all agents involved in the same interaction by an asymmetric convolutional network. 
	Based on the global information and the hierarchical attention, the final interaction is obtained and fed into a Graph Convolutional Network~(GCN)~\cite{mohamed2020social}
	which is followed by a Temporal Convolutional Network~(TCN)~\cite{bai2018empirical}, to estimate the parameters of Gaussian Mixed Model~(GMM)~\cite{reynolds2009gaussian}.
	
	
	Experimental results on multiple benchmark datasets demonstrate significant performance improvement of our method over the state-of-the-art methods. The visualization shows our method can learn the interaction among heterogeneous agents well. The code will be published upon acceptance.

	In summary, the key contributions of this paper include:
	\begin{itemize}[nosep]
    \item[$\bullet$] We propose to model the interaction among heterogeneous agents to improve the trajectory prediction; 
    \item[$\bullet$] We present an Unlimited Neighborhood Interaction for modeling the interaction among the agents involved in the same interaction simultaneously;
    \item[$\bullet$] We present a Hierarchical Graph Attention module for enhancing the agent-to-agent interaction based on category-to-category interaction.
    \end{itemize}
    
\section{Related Works}

\begin{figure*}[t]
    \centering
    \includegraphics[width=1.0\textwidth]{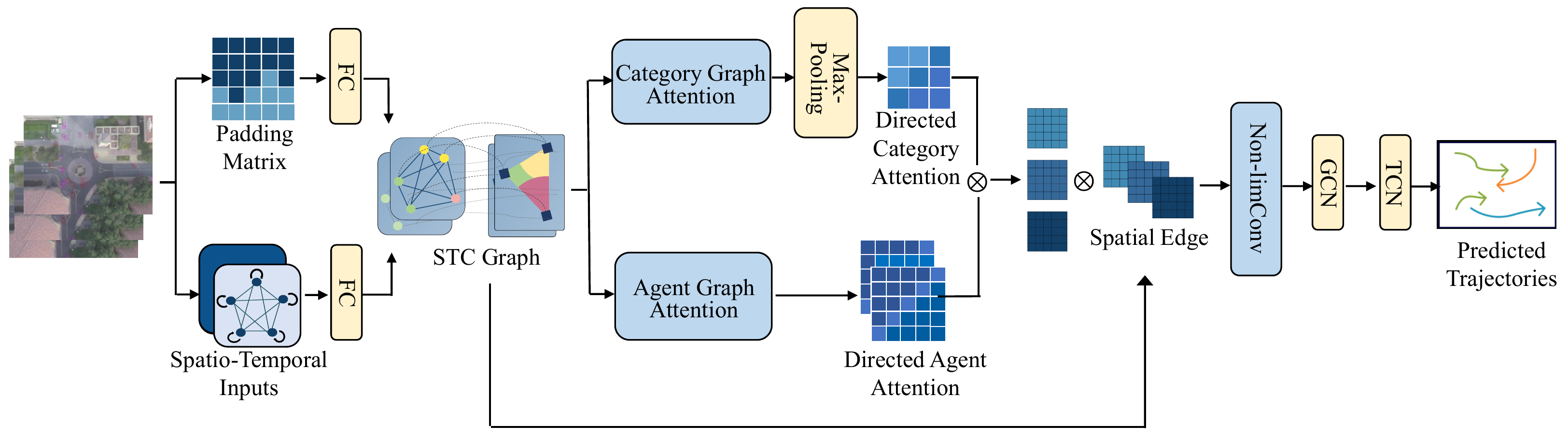}
    \caption{Framework of our UNIN. The trajectories are reformed as spatiao-temporal and category inputs, and a spatio-temporal-category graph (STC graph) is composed. Hierarchical Attention learns directed category attention representing the category interactions, and directed agent attention representing the agent interactions from the STC graph. Collective interactions are captured by subsequent Unlimited Neighborhood Interaction with the asymmetric attention matrix, and then fed into a spatio-temporal graph convolutional network and temporal convolutional network to estimate the parameters of the Gaussian Mixture Model, from which the future trajectories are predicted.}
    \label{fig:main}
\end{figure*}

Trajectory prediction mainly involves homogeneous and heterogeneous trajectory prediction in real scenarios. \emph{Homogeneous} trajectory prediction predicts future trajectories under the same category~(\eg, only pedestrians). On the contrary, \emph{heterogeneous} predicts future trajectories under different categories~(\eg, pedestrians, cars and bikes).
	
\subsection{Homogeneous Trajectory Prediction}

    Prior to the prevalence of deep learning, there are classical methods~\cite{tay2008modelling,treuille2006continuum,wang2007gaussian}, including Social Force models~\cite{helbing1995social}, Gaussian Process regression models~\cite{tay2008modelling}, dynamic Bayesian models~\cite{wang2008unsupervised} and hidden Markov models~\cite{surana2014bayesian},
    which are limited by hard-to-design hand-crafted features.
	
    Thanks to the representational power of deep neural networks, trajectory prediction is recently dominated by deep learning based methods, such as Recurrent Neural Networks (RNNs)~\cite{alahi2016social}, Generative Adversarial Networks (GANs)~\cite{gupta2018social}, Graph Convolutional Networks (GCNs)~\cite{mohamed2020social,Shi_2021_CVPR} and Transformers~\cite{yu2020spatio}. S-LSTM~\cite{alahi2016social} aggregates the interaction information through a pooling mechanism. S-GAN~\cite{gupta2018social} predicts multiple socially acceptable trajectories using GANs. Later works measure the influence of interaction by attention mechanism. S-BiGAT~\cite{kosaraju2019social} uses Graph Attention Networks~\cite{velivckovic2017graph} to model the interactions between pedestrians. STAR~\cite{yu2020spatio} separately models spatial interaction and temporal continuity through Transformer~\cite{vaswani2017attention} architecture on graph.
	
	Since physical constraints in the scenario and human states are the predominant factors in trajectory prediction~\cite{sadeghian2019sophie} under certain circumstances, extensive studies focus on the role of physical information recently~\cite{casas2020implicit,tao2020dynamic,liang2020garden}. Sophie~\cite{sadeghian2019sophie} leverages both physical and social information to predict pedestrian trajectory. Notably, CVM~\cite{hasan2018seeing} takes pedestrians’ velocity and direction into account rather than semantic environment. ECTP~\cite{mangalam2020not} infers trajectory endpoints first as additional information to assist pedestrians' planning path. Different from pedestrian trajectory prediction, vehicle trajectory prediction methods can take advantage of more sensors and semantic environments, such as 3D point cloud and lane line~\cite{liang2020learning,xie2017vehicle}.  

\subsection{Heterogeneous Trajectory Prediction}
	
	 Homogeneous traffic agents like pedestrians, vehicles follow different social conventions, and thus the homogeneous trajectory prediction methods cannot simultaneously model the interaction of all agents of different categories in the same scene and predict accurately.

	Heterogeneous trajectory prediction in the real traffic scenes gradually attracted more research interest. JPKT~\cite{bi2019joint} treats vehicles as rigid particles, where non-particle objects are subject to kinematics, and model vehicles and pedestrians with separate LSTMs. DATF~\cite{park2020diverse} models agent-to-agent and agent-to-scene interactions and proposes a new approach to estimate trajectory distribution. In brief, these methods focus on different behavior patterns of heterogeneous traffic agents, and the influence of the semantic environment.
	
	While previous works ignore the interaction at the categorical granularity and unlimited interactions among agents, we model the interaction among all agents. In our method, physical constraints are implicitly learned through observed trajectories without environmental semantics as prior.

\subsection{Graph Neural Network}
	
	Graph neural network~(GNN)~\cite{scarselli2008graph} extends the neural network to process data without nature order. GNN learns a state vector embedding containing information about every node and its corresponding neighbors. In order to gather information from neighbor nodes and their edges and enrich the representation of GNN, extensive works~\cite{kipf2016semi,hamilton2017inductive,li2015gated,velivckovic2017graph} study more complex graph structures. GCN~\cite{kipf2016semi} and Graph Sage~\cite{hamilton2017inductive} use spectral and spatial convolutional aggregation respectively, in which spectral convolution utilizes Fourier frequency domain to calculate graph Laplacian eigenvalue decomposition and spatial convolution operates on adjacent neighbor nodes in the spatial domain. GGNN~\cite{li2015gated} proposes a gated graph neural network to improve long-term information dissemination. GAT~\cite{velivckovic2017graph} introduces the attention mechanism to acquire the hidden state of the node by adding attention to its neighbor nodes. Highway GCN~\cite{rahimi2018semi} leverages skip connection to avoid introducing more noise from superimposing~\cite{li2018deeper} on the network layer. 
	
	The previous trajectory prediction methods, \eg, GCN and GAT, lack of a clear and proper distinction between heterogeneous nodes and homogeneous nodes, while our method takes large scale heterogeneous graph into account. 
	In addition, most existing graph neural networks group heterogeneous nodes into a subgraph, which suffer from data imbalance and ineffective global information aggregation. In contrast, we utilize hierarchical graph attention to aggregate the information of large-scale heterogeneous nodes.

\section{Our Method}

	In this section, we introduce our proposed UNIN, which aims to model interactions of heterogeneous traffic agents under the guidance of unlimited neighborhood interaction.
	Given a succession of video frames of traffic scenarios over time 
	$t \in \{1,2,\ldots,T_\text{obs} \}$, there are $C$ categories with $N$ agents.
	The goal of trajectory prediction is to predict the location of each traffic agent $i \in \{1,\dots,N\}$  within a future time horizon $ t \in \{T_{obs+1},T_{obs+2},\ldots,T_\text{pred}\} $. For a traffic agent $i \in \{1,\dots,N\}$ of category $ c \in \{1,2,\dots, C\} $ , it is denoted as $ V_t^i = \left( x_t^i, y_t^i, c\right) $ , where $\left(x_t^i, y_t^i\right)$ is the location coordinate of traffic agent $i$ at time step $t \in \{1,2,\ldots,T_\text{pred}\} $.

	As discussed, the interactions in previous works are only considered between two traffic agents or in a local area, while an unlimited number of other agents may be simultaneously involved in an interaction regardless of their category.
	Additionally, most of the existing works neglect people’s diverse reactions to heterogeneous agents, which is spontaneous in real traffic scenarios, is under-explored. 
	To mitigate these limitations, we propose the Unlimited Neighborhood Interaction to capture the impact that all agents experience at the same time, and a hierarchical attention module to model heterogeneous interactions among traffic agents of different categories.

	The overall framework of UNIN is illustrated in Figure~\ref{fig:main}. To aggregate information of agents involved in the same interaction, an interaction graph is built first to gather global interaction information. 
	Subsequently, the Hierarchical Attention Module is used to obtain the category-category interaction and agent-agent interaction based on the global interaction information.
	Next, we introduce the Unlimited Neighborhood Module directly modeling interactions 
	by pooling features among unlimited neighborhood agents.
	Finally, a heterogeneous graph convolution network and a temporal convolutional network are used to predict the parameters of a Gaussian Mixture Model for trajectory prediction. 
	
\subsection{Heterogeneous Graph Construction}
    
	There are agents in multiple categories in heterogeneous trajectory prediction, and thus
	we build a spatio-temporal-category graph $\mathcal{G}_{stc}$ to model them altogether, as shown in Figure~\ref{fig:main}, where every agent is regarded as a node and the interactions among agents are regarded as edges. To enhance the representations of category-category interaction, we also regard all agents with the same category as a category node:
	\begin{equation}
	    \mathcal{G}_{stc} = ( V_t^i, E_t^i, E_t^{ij}, S_t^c, D_t^{c_1,c_2}, D_t^c),
	\end{equation}
	where $i \in \{1,...,N\}, t \in \{1,\dots,T_{pred}\}, c=\{1,\dots,C\}$ represent the index of node, time step and category, respectively. $V_t^i = \{ ( x_t^i, y_t^i, c) \}$ represents the node $i$ with category $c$ at the time step $t$. $ E_t^i = \{ ( V_t^i, V_{t+1}^i) \}$ is a temporal edge connecting  node $V_t^i$ and $V_{t+1}^i$.  $E_t^{i,j} = \{ ( V_t^i, V_{t}^j) \}$ is a spatial edge connecting node $V_t^i$ and $V_t^j$. $S_t^c = \{V_t^i \mid \forall i \in \{1, \dots, N\}\}$ is the category node with category $c$ at time step $t$ generated by the concatenation of all agents with category $c$ at time step $t$, which represents the embedding, \ie,  the concatenation of the agents of category c. We consider the difference of various categories of agents, and project them with a common transformation matrix. Then we concatenate the agent features of the same category. $ D_t^{c_1, c_2} = \{ (S_t^{c_1}, S_t^{c_2}) \mid c_1, c_2 \in \{1,\dots,C\} \}$ is the spatial category edges connecting category node $S_t^{c_1}$ and $S_t^{c_2}$ at time step $t$. $ D_t^c = \{\left(S_t^c, V_t^i\right) \} $ is the spatial category-agent edges connecting category node $S_t^c$ and each spatial node $V_t^i$ belonged to category $c$.
	
	The built spatio-temporal-category graph $\mathcal{G}_{stc}$ includes not only the information of each agent, but also the information of each category. Therefore, we can leverage $\mathcal{G}_{stc}$ to build category-to-category and agent-to-agent interaction.

\subsection{Hierarchical Graph Attention}

The interaction among agents is an essential factor for trajectory prediction.
Especially, the heterogeneous interaction is more complex due to diverse object categories compared with homogeneous interaction~\cite{niu2017Hierarchical}. In traffic scenarios, 
traffic agents (pedestrians, drivers, bikers, etc.) tend to react differently according to the categories of agents they encounter because of the difference in social habits and experiences. Hence, the interaction between categories (\ie, category-category interaction) is also an important factor affecting agent's trajectories.
	

In order to model the interaction among agents with multiple categories, we propose a Hierarchical Graph Attention module. It models the category-category interaction first, based on which the agent-agent interaction is modeled.
	
	
\textbf{Category-Category Interaction}. To build the interaction among categories, we obtain the category features of each category first on our built spatio-temporal-category graph, based on which the category-wise interaction weights are obtained through pooling operation.

In light of the imbalanced amount of agents in different scenarios, we employ a padding operation to align them to the same amount. Then, the embeddings $h_t^c$ of each category are obtained by a linear projection, \ie, 
\begin{equation}
h_t^c = \phi\left(W_e, \Theta \left(S_t^c\right)\right), 
\end{equation}
where $\phi\left(\cdot, \cdot\right)$ denotes linear projection, $S_t^c$ is the category node with category $c$ at time step $t$, $h_t^c$ is the embedding of category $c$ at time step $t$, $\Theta$ is the padding operation,  and $W_e$ is the learnable weight of linear projection. The padding size equals to the largest number of nodes in the scenario for efficient computation. Padded convolution is also flexible for arbitrary number of agents, as convolution on 0 will not change the result. 

After acquiring the embeddings of each category, the embeddings of any two categories are concatenated to obtain fused embeddings. 
Subsequently, the category-category attention scores $A_t$ are generated by graph attention mechanism~\cite{velickovic2018graph}, as follows:
	 \begin{equation} \label{eq:gat}
	     A_{t}^{c_1,c_2} =\delta\left( \mu_{c} \cdot (h_t^{c_1}~\|~h_t^{c_2}) \right),
	 \end{equation}
	 where $A_{t}^{c_1, c_2}$ is the attention score vector of category $c_1$ to $c_2$ at time step $t$, $\mu_{c}$ denotes a learnable attention weight vector of category $c$ used to adjust the weights among categories,  $\delta(\cdot)$ denotes a non-linear activation function.
	 
	 The attention score vector $A_{t}^{c_1, c_2}$ measures the interaction of one category to other categories.
     The category-category interaction aims to assist agent-agent interaction, and thus we only acquire an importance factor by pooling operation for each attention score vector $A_{t}^{c_1, c_2}$.  We employ the max pooling($\Upsilon$) to choose the biggest value in $A_{t}^{c_1, c_2}$ as the importance factor
      $a_{t}^{c_1, c_2}$,
     \ie,
	 %
	 \begin{equation}
	 a_{t}^{c_1, c_2}=\Upsilon \left(A_{t}^{c_1, c_2}\right).
	 \end{equation}

     After acquiring the importance factor between any two categories, the final category-category interaction $\text{CI}_{t}^{c_1, c_2}$ is obtained by normalizing all the importance factors (which the number of categories is n):   
	 \begin{equation}
	     \text{CI}_{t}^{c_1, c_2}=\frac{\exp \left( {a_t^{c_1, c_2}}\right)}{\sum_{i, j \in n} \exp \left( {a_{t}}^{i, j}\right)}.
	 \end{equation}
	 The weights of spatial category edges $ D_t^{c1, c2} $ represent the category-category interaction, and thus we assign value to $D_t^{c1, c2}$ by the obtained interaction values.

\textbf{Agent-Agent Interaction.} Some related works~\cite{mohamed2020social} indicate the relative distance between agents is essential in some special scenarios. Therefore, we obtain the agent-agent interaction by a combination of learning-based method and distance-based method.

It is intuitive to define a weight that grows for approaching agents based on an assumption that agents are more susceptible to closer ones. Meanwhile, the attention mechanism ensures that far interacting agents can also be recognized by the model.

The distance-based method initializes the spatial edge $E_t$ with the relative distance between the corresponding agents. Then, the normalized interaction matrices 
${R}_t$ is obtained by Laplace Transform~\cite{Masuda2017A} as follows:

\begin{equation}
	 \begin{split}
	  	 &{E_t^{i,j}}=
	    \left\{\begin{array}{ll}1 /\|p_{t}^{i}-p_{t}^{j}\|_{2}, & \|p_{t}^{i}-p_{t}^{j}\|_{2} \neq 0 \\ 0, & \text { Otherwise}\end{array}\right.,\\
	    &R_{t}=\Lambda_{t}^{-\frac{1}{2}} \hat{E}_{t} \Lambda_{t}^{-\frac{1}{2}},
	 \end{split}
	 \end{equation}
where $p_{t}^{i}, p_{t}^{j}$ is the location coordinates for agent $i,j$ at time step $t$, $\hat{E}_{t} = E_{t} + I$, and $\Lambda_{t}$ is the diagonal node degree matrix of ${E}_{t}$.

	 

For the learning-based method, we need to fuse the features of all agents. Fortunately, 
the learned attention score vector $A_t$ shown in Equation~\ref{eq:gat} already includes the required information, and thus we directly employ the learned $A_t$ to obtain the agent-agent interaction $\text{ATT}_{t}$, \ie, 
	  \begin{equation}
	  	\text{ATT}_{t}=R_{t} \otimes A_t,
	 \end{equation}
where operator $\otimes$ denotes dot-product operation.

\begin{table*}[t]
\scriptsize
\centering
\setlength{\tabcolsep}{4.8mm}{
\begin{tabular}{c|cc|cc|cc|p{7mm}<{\centering}p{7mm}<{\centering}}
\toprule
    \multirow{2}*{Models}  & \multicolumn{2}{c}{Argoverse} & \multicolumn{2}{c}{nuScenes}&\multicolumn{2}{c}{Avg}&\multicolumn{2}{c}{Apolloscape} \cr \cmidrule(lr){2-9} & ADE & FDE & ADE & FDE & ADE & FDE & WADE & WFDE \\     
\midrule
\text{S-LSTM}~\cite{alahi2016social}   &  1.385  & 2.567    & 1.390 & 2.676&1.388 &2.622 &  1.89 & 3.40  \\
\text{DESIRE*}~\cite{lee2017desire}  &  0.896  & 1.453  & 1.079  & 1.844  &0.988 &1.649  &  - & - \\
\text{R2P2-MA}~\cite{rhinehart2020r2p2}  &1.108 & 1.771 & 1.179  & 2.194   &1.144 &1.983 &  - & - \\ 
\text{CAM}~\cite{park2020diverse}  &   1.131   & 2.504   & 1.124   & 2.318  &1.128 &2.411  & - & -  \\
\text{MFP}~\cite{tang2019multiple}  &  1.399   & 2.684  & 1.301  &2.740  & 1.350  &2.712 &  - & -  \\
\text{MATFD}~\cite{zhao2019multi}  &  1.344 & 2.484 & 1.261 & 2.538      &1.303 &2.511    &  - & -  \\
\text{MATFG*}~\cite{zhao2019multi}  &  1.261 & 2.313 & 1.053 & 2.126      &1.157 &2.220   &  - & -  \\
\text{STGCNN}~\cite{mohamed2020social}  &  1.305 & 2.344 & 1.274 & 2.198   &1.289 &2.371 & - & -   \\
\text{StarNet}~\cite{zhu2019starnet}   &  - & - & - & -      &- &-     &1.343 &2.498  \\
\text{TPNet}~\cite{fang2020tpnet} &  - & - & - & -      &- &-     &1.281 &1.910   \\
\midrule
\textbf{NLNI (Ours)}    &  \textbf{0.792}   & \textbf{1.256}   & \textbf{1.049}   & \textbf{1.521} &\textbf{0.921} &\textbf{1.388} &\textbf{1.094} &\textbf{1.545} \\
\bottomrule
\end{tabular}
}
\caption{Comparison with other methods on dataset Argoverse, nuScenes and Apollscape in ADE and FDE metrics (the lower the better). All methods observe $2$ seconds and predict the next $3$ seconds of trajectories and are evaluated on validation set and test set. Note that the Apolloscape dataset and other validation sets use weighted ADE and FDE metric, \ie, the weights of vehicles, pedestrians and cyclists are assigned as 0.20, 0.58 and 0.22, respectively. The methods marked by ``*`` use additional scene context. Our UNIN significantly outperforms the state-of-the-art works.
}
\label{tab:performance}
\end{table*}

\begin{table*}[t]
\scriptsize
\centering
\resizebox{1.0\linewidth}{!}{
\setlength{\tabcolsep}{0.5mm}
\begin{tabular}{c|c|c|c|c|c|c|c|c|c|c|c}
\toprule
\multirow{2}*{Datasets}& \multicolumn{11}{c}{Models} \cr \cmidrule(lr){2-12}
& S-LSTM~\cite{alahi2016social} & MATF~\cite{zhao2019multi} & DESIRE~\cite{lee2017desire} & NRI~\cite{kipf2018neural} & S-GAN~\cite{gupta2018social} & SOPHIE~\cite{sadeghian2019sophie} & Traject++~\cite{salzmann2020trajectron++} & STGCN~\cite{mohamed2020social} & SIMAUG*~\cite{mohamed2020social} & STGAT~\cite{kosaraju2019social}
& \textbf{Ours}    \\
\midrule
\text{\text{SDD}}  & 31.2 / 57 & 22.6 / 33.5 & 19.3 / 34.1 & 25.6 / 40.3 & 27.3 / 41.4 & 16.3 / 29.4 & 19.3 / 32.7 & 20.6 / 33.1 & \textbf{15.7} / 30.2 & 18.8 / 31.3 & \textbf{15.9} / \textbf{26.3} \\
\bottomrule
\end{tabular}
}
\caption{Comparison with the previous approaches on the SDD benchmark dataset, which mainly contains the trajectories of pedestrian. The performance is evaluated in ADE/FDE metrics (the lower the better). The approach marked by ``*`` uses additional simulation data.}
\label{tab:performance2}
\end{table*}


\subsection{Unlimited Neighborhood Interaction}
In a real traffic scenario, interactions differ among the uncertain numbers of agents, \ie, an agent could respond differently as the number of interacted agents varies.
However, the existing graph attention mechanism~\cite{wang2019heterogeneous} only computes the interaction between pair-wise agents because the inner-product is operated only between two vectors once. And graph convolutional network with many layers suffers from over-smoothing~\cite{1NEURIPS2019_73bf6c41, 2018arXiv181005997K}. According to our observation, GCN with one or two layers is optimal for our task.
 Hence, the learned agent-agent interaction $ATT_t$ can not adaptively capture the interaction among the uncertain number of agents.

To mitigate this, we propose the Unlimited Neighborhood Interaction module to capture the information of all agents involved in a same interaction simultaneously. Note that all agents involved in an interaction are called ``unlimited neighborhood'', regardless of the numbers of agents.
In particular, we employ an asymmetric convolution to obtain and aggregate the global interaction information on $ATT_t$, \ie,
	   \begin{equation}
	       h_t = \delta (\operatorname{Conv1D} (\text{ATT}_{t})),
	   \end{equation}
where $\delta$ is the non-linear activation function, and we use padding operation to ensure the output size the same as the input size.

The asymmetric convolution is computed repeatedly and thus the global spatial interaction information can be aggregated, meaning that all agents involved in an interaction are considered, regardless of the number of the agents. Because small asymmetrical kernel with padding captures implicit interactions, which are not limited by the number of agents or the range of area. It ensures any number of agents in a specific interaction can be considered, while a big symmetric kernel mixes different numbers/ranges of agents in different interactions.
	 
	  The final interaction $F_t$ is obtained through fusing unlimited neighborhood and category-category interaction:
	  \begin{equation}
	      F_t = \text{CI}_{t}^{c_1, c_2} \otimes h_t.
	  \end{equation}

\begin{table*}[t]
\scriptsize
\centering

\setlength{\tabcolsep}{5mm}{
\begin{tabular}{c|c|c|c|c|c}
\toprule 
\multirow{2}{*}{Dataset} & \multicolumn{2}{c|}{MLP} & \multicolumn{2}{c|}{CNN} & \multirow{2}{*}{NLIN (Ours)}\tabularnewline
\cline{2-5} 
 & w/o HGA & w/o UNI & w/o HGA & w/o UNI & \tabularnewline
\midrule 
Apolloscape & 1.460/1.794 & 1.576/1.843 & 1.837/2.014 & 1.792/1.955 & \textbf{1.094/1.545}\tabularnewline
\hline 
nuScenes & 1.613/1.969 & 1.547/1.728 & 1.763/1.982 & 1.701/1.934 & \textbf{1.049/1.521}\tabularnewline
\bottomrule 
\end{tabular}
}

\caption{The ablation study of each component (Using MLP/CNN to replace each component). UNIN (Ours) combines with each component.
}
\label{tab:ablation}
\end{table*}

\begin{table}[t]
\scriptsize
\centering
\label{tab:my_label}

\begin{tabular}{c|ccccc}
\toprule
UNIConv Size & 1 & 2 & 3 & 5 & 10\tabularnewline
\midrule
ADE & 1.179 & \textbf{0.921} & \textbf{0.998} & 1.247 & 2.691\tabularnewline
\hline 
FDE & 1.632 & \textbf{1.388} & \textbf{1.323} & 1.766 & 3.515\tabularnewline
\bottomrule
\end{tabular}

\caption{Ablation study of kernel size for Unlimited Neighborhood convolution. 
}
\label{tab:ablation2}
\end{table}
\subsection{Trajectory Prediction}
     After obtaining the final interaction $F_t$, we regard it as the adjacency matrix of the spatio-temporal-category graph and feed it in GCN, which is followed by a TCN to estimate the parameters of Gaussian Mixture Model. A residual connection is used in GCN, \ie,
     \begin{equation}
	 \begin{split}
	  H_t^{(l)}&=\delta (H_t^{(l-1)} + F_t^{(l)} \cdot \mathrm{Conv} (H_t^{(l-1)})), \\
	   HT &= \operatorname{TCN} (H_t), 
	 \end{split}
	 \end{equation}
    where $\delta$ is a non-linear activation function, $l$ is the index of layers of GCN, $H_t^0=V_t$ represents the node of the graph, and $HT$ is the output features of TCN.
	Thus we acquire the collective interaction information from both the space and the time information. 
	
	 
	 \textbf{Loss Function.}~Since the traffic agents of different categories have their own  unique movement pattern, \eg, a certain velocity range, front and rear distance to another object, 
     we assume that the trajectory coordinates $(x_t^i,y_t^i)$ of traffic agents $i$ follow a Gaussian Mixture Model~\cite{Dong2017Gaussian}. Hence, our model is trained by minimizing the negative log-likelihood loss as follows:
	 \begin{equation}
	     \mathrm{L^i}=-\sum_{t=T_{obs}+1}^{T_{pred}} \text{log} \sum_{k=1}^{K} \pi_{k} N\left((x_t^i,y_t^i) \mid \hat{\mu}_{n}^{t}, \hat{\sigma}_{n}^{t}, \hat{\rho}_{n}^{t} \right),
	 \end{equation}
where $\hat{\mu}_{n}^{t}$ is the mean, $\hat{\sigma}_{n}^{t}$ is the standard deviation, $\hat{\rho}_{n}^{t}$ is the correlation co-efficient, and $\pi_{k}$ is the weight factor of the $k$-th Gaussian distribution.

\section{Experiments}
	\textbf{Datasets.} 
	Some datasets focus on homogeneous trajectories, and contain fewer traffic scenes, \eg, ETH~\cite{pellegrini2009you} and UCY~\cite{lerner2007crowds}, which only label pedestrian trajectory within three scenes. However, there are often diverse categories in the real scenario, and thus we train and evaluate our model on more complex datasets, including Stanford Drone Dataset(SDD)~\cite{robicquet2016learning}, nuScenes~\cite{caesar2020nuscenes}, Argoverse~\cite{chang2019argoverse} and Apolloscape~\cite{ma2019trafficpredict}, which are widely used in heterogeneous trajectory prediction with diverse categories and rich traffic scenes.
	The SDD consists of eight unique scenes on the university campus, more than 100 static scenes, $19K$ traffic agents of $6$ categories, and approximately $40K$ interactions.  
	The nuScenes, Argoverse and Apolloscape are large-scale trajectory datasets for urban streets with dense traffic in highly complicated situations. Besides, trajectories in them are collected through an in-vehicle camera so that they have more different scenarios.
	
		We follow the existing works, observing 3.2 seconds of trajectories while predicting the next 4.8 seconds in Stanford Drone Dataset, and observing 2 seconds while predicting the next 3 seconds in nuScenes, Argoverse, and Apolloscae datasets.
	

	\textbf{Evaluation Metrics.}
	We follow existing works~\cite{yu2020spatio} and employ two common metrics to evaluate the performance: Average Displacement Error (ADE) and the Final Displacement Error (FDE), which are defined as follows: 
	
\begin{equation}
\begin{split}
&\mathrm{ADE}=\frac{\sum_{n \in N} \sum_{t \in T_{p}}\left\|\hat{p}_{\mathrm{t}}^{n}-p_{\mathrm{t}}^{n}\right\|_{2}}{N \times T_{p}},\\ 
&\mathrm{FDE}=\frac{\sum_{n \in N}\left\|\hat{p}_{T}^{n}-p_{T}^{n}\right\|_{2}}{N \times T_{p}},
\end{split}
\end{equation}
where ADE measures the average L2 distance between ground truth and our predicted future positions over all time steps, while FDE measures the L2 distance between our predicted final destination and the true final destination. 

\subsection{Implementation Details}
    In the Hierarchical Attention Module, the embedding dimension of one category is set to 8 and the output size after padding is equal to the largest number of nodes in the scenario.
    In the Unlimited Neighborhood Module, the kernel-size $k$ of the convolution(UNIConv) is fixed at 3.
    We train our model with SGD, and the learning rate is set to 0.005, which decays by a factor $0.2$ after every $10$ epochs. The weighted factor of GMM loss is acquired from the Hierarchical Attention Module and the approximate ratio of the categories in scenes. 
    We train our model on an RTX2080Ti GPU for up to 50 epochs. 
    And we use a dataset split of 60\%, 20\%, 20\% for training, validation and testing, respectively. The complete code will be published once upon acceptance.   


\begin{figure*}[t]
    \centering
    \includegraphics[width=1.0\textwidth]{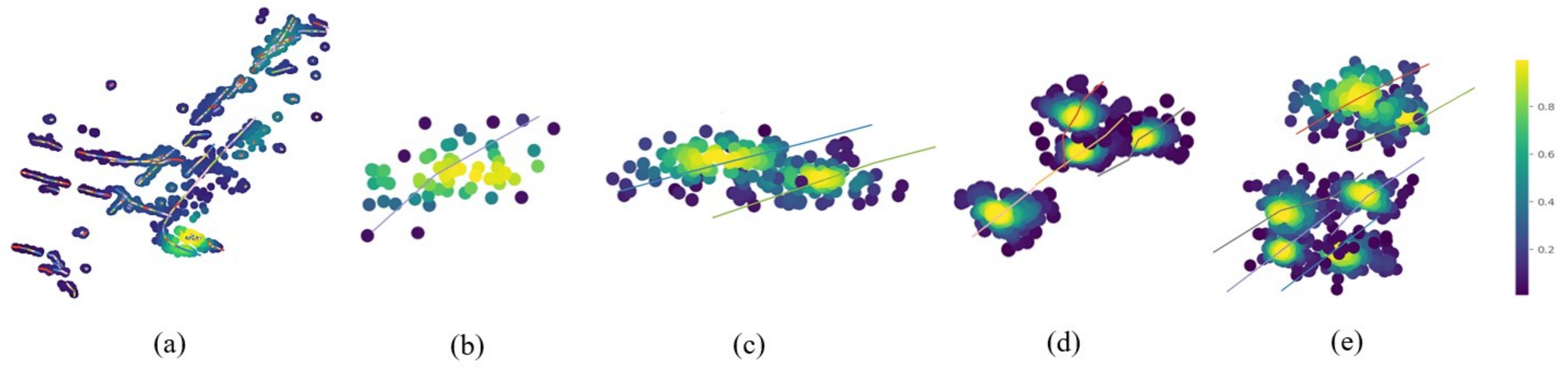}
    \caption{\textbf{Visualization of predicted trajectory distribution.} 
    Each line represents the ground-truth trajectory of an agent. 
    The colored dots represent our predicted trajectory distribution, and different colors represent different densities of our predicted distribution, where yellow represents the most likely trajectory distribution.
    (a) shows the overall trajectories in the whole scene at all of the time instants. 
    (b) shows that we successfully predict a turning agent.
    (c) shows that we successfully predict two agents going in parallel to the same direction.
    (d) shows that we successfully predict two agents separated by another interacting and avoiding each other.
    (e) shows that we successfully predict the possible trajectory of a group of agents after collective interaction.
    All results are randomly sampled from the nuScenes dataset.
    }
    \label{fig:visual}
\end{figure*}

\subsection{Quantitative Evaluation}
	Table~\ref{tab:performance} and Table~\ref{tab:performance2} show the comparison of our method against state-of-the-art approaches, including Social LSTM~\cite{alahi2016social}, Social GAN~\cite{gupta2018social}, STGAT~\cite{kosaraju2019social}, Social STGCNN~\cite{mohamed2020social}, Trajectron++~\cite{salzmann2020trajectron++}, NRI~\cite{kipf2018neural}, SoPhie~\cite{sadeghian2019sophie}, MATF~\cite{zhao2019multi}, DESIRE~\cite{lee2017desire},  SimAug~\cite{liang2020simaug}, P2P2-MA~\cite{rhinehart2020r2p2}, CAM~\cite{park2020diverse}, MFP~\cite{tang2019multiple}, StarNet~\cite{zhu2019starnet} and TPNet~\cite{fang2020tpnet}.
	%
	Overall, our method significantly outperforms all compared methods on all datasets according to the tables.
	 Particularly, our UNIN surpasses the DESIRE (the second best) by $2.7\%$ on average in ADE and  $13.85\%$ on average in FDE for nuScenes, Argoverse and Apolloscape.
	 Meanwhile,  our method achieves a performance improvement by $10.5\%$ on average in FDE for SDD dataset. 
	The underlying reason is that our method can model the collective interaction among the agents involved in the same interaction simultaneously. Meanwhile, the Hierarchical Attention enhances the agent-agent interaction based on category-category interaction.

\textbf{nuScenes, Argoverse and Apolloscape.} Our UNIN outperforms all the competing methods on the three datasets. 
	%
    The nuScense, Argoverse and Apolloscape are multi-category mixed datasets with a majority of vehicles.
    Compared with the RNN-based method, such as S-LSTM~\cite{alahi2016social}, our method surpasses it by $42.8\% /51.1\% $ in FDE/ADE metrics.  We speculate that S-LSTM employs a pooling mechanism to aggregate local agents' states, while it does not take the long-range interaction into account.
    In addition, our method also  outperforms the Graph-based methods, \eg, S-STGCNN~\cite{mohamed2020social}, by $28.5\% /41.3\% $ in FDE/ADE metrics. We speculate it takes the long-range interaction into account but the interactions are only modeled between pairwise agents.
    Interestingly, our method outperforms the methods employed scene context, such as DESIRE~\cite{lee2017desire} and MATFG~\cite{zhao2019multi}. Both of them employ a LSTM to model each agent and fuse the interaction with in a local area, while our model considers the unlimited neighborhood, which is not limited by the number of agents and the range of interaction. Thus, our method can capture more global and local detail information to improve the accuracy of future trajectories.

	\textbf{Stanford Drone Dataset.}
	Stanford Drone Dataset(SDD) is a multi-category mixed dataset including pedestrians, bicyclists, skateboarders, carts, cars and buses with a majority of pedestrians.
	Our method outperforms the methods modeling the interaction in a local area, such as S-LSTM~\cite{alahi2016social}~(ours achieves $49\% /52\% $ better on average in ADE/FDE) and S-GAN~\cite{gupta2018social}~(ours achieves $41.7\% /36.5\% $ better on average in FDE/ADE). We speculate the reason is they employ a pooling mechanism to aggregate the local agent's interaction states, while our method employs an unlimited interaction capable of capturing the information of flexible interactions. 
	In addition, our method is better than the graph-based methods, such as STGCNN~\cite{mohamed2020social}, by $22.8\% /20.5\% $ average. 
	Moreover, our method is slightly outperformed by SIMAUG~\cite{liang2020simaug} in ADE metric, possibly due to SIMAUG uses extra 3D simulation data for training, leading to more robust representations.
	We also evaluate the data efficiency and generalization ability of our model, please refer to supplemental material for detail.

	\subsection{Qualitative Evaluation}
	We further study the ability of our method to model interactions of large-scale traffic agents with multiple categories. As discussed previously, there are often interactions with large numbers of agents and uncertain distances between them in real traffic scenarios. And agents often adopt different strategies when interacting with different categories of traffic participants. We illustrate some qualitative evaluation results in Figure~\ref{fig:visual}. 
	Overall, our predicted trajectory distributions are in line with the ground truth trajectories.
    Result (a) is the long time trajectories from the beginning time instant to the last time instant, which demonstrates the great prediction accuracy achieved by our method. 
    Result (b) shows a single traffic agent that is turning. As expected, our model captures the agent's tendency of turning. 
    Result (c) shows our method successfully predicts the trajectory when two agents are going in parallel orienting towards the same direction, which means our method does not appear to be over-fitting. 
    In (d), two non-adjacent agents interact with each other rather than with another closest agent to them. Our method leverages the UNI to capture the long-range interaction, successfully predicting that relatively distant agents interacting and the subsequent trajectories.
    Result (e) shows the collective interaction involving a group of agents belonging to different categories. Our method successfully predicts the possible trajectory of them with a complex interaction. And our predicted trajectory distributions show that the agents of different categories react differently when interacting with a specific agent, which demonstrates the efficiency of our HGA. We also visualize the relation between category attention and agent attention in supplemental material.
	
	\subsection{Ablation Study}
	We study the contribution of each component in our model as shown in Table~\ref{tab:ablation}. In addition, we set different values of kernel size of Unlimited Neighborhood Interaction to find the empirical optimal value, as shown in Table~\ref{tab:ablation2}.
	
	
	\textbf{Contribution of Each Component.} As illustrated in Table~\ref{tab:ablation}, we evaluate two variants of our method: \textbf{(1)} UNIN w/o HGA, which means the category-to-category attention is replaced with CNN/MLP and only the agents-to-agents interaction is kept; \textbf{(2)} UNIN w/o UNI, which means the unlimited neighborhood interaction is replaced with CNN/MLP.
	According to the results, removing any component will lead to a large performance drop. Particularly, the results of UNIN w/o HGA show a performance reduction by $28.3\% / 16.4\% $  in ADE/FDE metrics, reflecting the effectiveness of hierarchical attention.
	The results of UNIN w/o UNI shows a performance degradation by $30.1\% /14.2\% $ in ADE/FDE metrics, which validates the contribution of unlimited neighborhood interaction. 

	\textbf{Optimal Kernel Size.} As shown in Table~\ref{tab:ablation2}, the optimal value of the kernel size of Unlimited Neighborhood Interaction convolution is $2$ in ADE metric, and $3$ in FDE metric. 
 From the table, a larger kernel size is unhelpful. The convolution with kernel size 2 and 3 are the best performing settings to capture the relation among group agents. 

\section{Conclusion}
    To capture the interaction information with varying numbers of agents from an uncertain distance, we present an Unlimited Neighborhood Interaction Network to predict trajectories in multiple categories.
    An Unlimited Neighborhood Interaction Module generates the interaction with all of the agents involved in the interaction simultaneously. 
    A Hierarchical Graph Attention module is designed to acquire the category-to-category interaction and agent-to-agent interaction, where the former one  is used to enhance the representation of agent-to-agent interaction.
    Extensive quantitative evaluations show our method achieves state-of-the-art performance, even outperforming methods leveraging additional scene context.
    Qualitative evaluations illustrate the advantage of our method when predicting heterogeneous trajectories in dense and complex traffic scenarios.
\section*{Acknowledgment}
    This work was supported partly by National Key R\&D Program of China Grant 2018AAA0101400, NSFC Grants 62088102, 61976171, and 61773312, the General Program of China Postdoctoral Science Foundation under Grant No. 2020M683490, Young Elite Scientists Sponsorship Program by CAST Grant 2018QNRC001, and the Youth program of Shanxi Natural Science Foundation under Grant No. 2021JQ-054.



{\small
\bibliographystyle{ieee_fullname}
\bibliography{main}
}
\end{document}